%% file: main2.tex
\documentclass[a4paper,12pt]{article}
\usepackage[utf8]{inputenc}
\usepackage[T2A, T1]{fontenc}

\usepackage{setspace,amsmath}
\usepackage{amssymb} 
\usepackage{dsfont}

\usepackage{systeme} 
\usepackage{mathtools} 
\usepackage{array} 
\usepackage{makecell} 
\usepackage{multirow} 
\usepackage{longtable}
\usepackage{subfiles}
\usepackage{hyperref}
\hypersetup{pdfstartview=FitH,  linkcolor=blue, urlcolor=blue, colorlinks=true}
\usepackage{framed} 
\usepackage{graphicx}
\usepackage{caption}
\usepackage{subcaption} 
\usepackage{color}
\usepackage{chngcntr} 
\usepackage{authblk} 
\usepackage{appendix}  

\usepackage{float} 
\floatstyle{plaintop} 
\usepackage{enumitem} 
\usepackage{booktabs} 

\usepackage[
backend=biber,
sorting=none, block=space, url=false]{biblatex} 
\graphicspath{{./img/}}

\newcommand{\defeq}{\overset{def}{=}}

\usepackage[left=30mm, top=20mm, right=30mm, bottom=20mm, nohead, footskip=10mm]{geometry}

\begin{document}
	
	\title{Towards a Relationship-Aware Transformer for Tabular Data}
	
	\author[1]{Andrei V. Konstantinov}
	\author[1]{Valerii A. Zuev}
	\author[1]{Lev V. Utkin}
	\affil[1]{\small Peter the Great St.Petersburg Polytechnic University \normalsize \\
		
		\texttt{andrue.konst@gmail.com, zuev\_va@spbstu.ru,  lev.utkin@gmail.com}}
	
	\date{}
	\maketitle
	
\section*{Abstract}

Deep learning models for tabular data typically do not allow for imposing a graph of external dependencies between samples, which can be useful for accounting for relatedness in tasks such as treatment effect estimation. Graph neural networks only consider adjacent nodes, making them difficult to apply to sparse graphs.
This paper proposes several solutions based on a modified attention mechanism, which accounts for possible relationships between data points by adding a term to the attention matrix.
Our models are compared with each other and the gradient boosting decision trees in a regression task on synthetic and real-world datasets, as well as in a treatment effect estimation task on the IHDP dataset.

\section{Introduction} \label{section:intro}

The gold standard for drug testing is the double-blind randomized controlled trial, which purpose is to estimate the average effect of a treatment across the population.
However, personalized healthcare requires individual treatment effect (ITE) estimation based on the patient's characteristics \cite{shalit_estimatingTarNet_2017, bica_realWorld_toTE_2021, angus_heterogeneity_2021}.
In this approach, a patient is represented by a set of features such as age, gender, smoking status, blood oxygen level, and so on.
This allows for estimating the potential impact of treatment conditional on the patient's health record.
In some cases, a therapy can be selected by trial and error,
but often the costs of inappropriate or suboptimal treatment are very high, such as with surgical intervention.
Therefore, predicting the success of treatment taking into account the individual characteristics of the patient is of great importance.

Usually the patient's condition after the treatment (or in the absence of treatment) is represented by a single continuous variable -- \textit{outcome}, and the task is to predict the difference between two outcomes.
The higher the difference, the more effective the treatment.
No patient can simultaneously be treated and untreated, making it impossible to create a real-world benchmark dataset for treatment effect estimation.
In this work, we compare the algorithms on a widely used semi-synthetic IHDP (Infant Health Development Program) dataset introduced by Louizos et al. \cite{louizos_cevae_2017}.

Treatment effect estimation has been explored in statistics, traditional machine learning, and deep learning \cite{chu_causal_2023, curth_using_2023}.
One can fit two different estimators for two outcomes (factual and counterfactual), but this might lead to a poor performance if the dataset is imbalanced or if treatment assignment is not random.
In 2017, Shalit et al. presented TarNet (Treatment-Agnostic Representation Network) \cite{shalit_estimatingTarNet_2017}
for binary treatment effect estimation. This method two neural networks with shared layers jointly trained to estimate factual (observed in experiment) and counterfactual outcomes.
In 2019, Shi et al. proposed training a single neural network called DragonNet for estimating the treatment effect (the difference between outcome when patient is treated and untreated), then removing the final layer of the network and training two distinct layers to estimate two outcomes \cite{shi_adapting_2019}.
Also, there are frameworks referred to as S-learner, T-learner and X-learner, that make it possible to build an efficient treatment effect estimator based on any regression model \cite{kunzel_metalearners_2019}.

As shown in multiple studies, genotype-defined traits may significantly affect the treatment outcome.
One example is ethnicity  \cite{greenbaum_genderEthnicityDrugTherapy_2015, arundell_differencesPsychoEthnicityGender_2024, cerdena_raceConscious_2020}, e.g. ethnic minorities seem to benefit more from psychotherapy if the therapist belongs to the same ethnic group.
Genotype is either unknown or represented by a high-dimensional vector, and therefore difficult to be incorporated into a feature vector.
However, we often possess an implicit knowledge about genotype, e.g. ethnicity or family ties between individuals; in our work, we refer to such information as \textit{relatedness}, or a graph of relationships.
To our knowledge, the problem of treatment effect estimation on a set of patients with known relatedness has never been previously addressed.

Mathematical constructs such as Nadaraya-Watson regression and the attention mechanism use the distance between all pairs of data points (defined by some proximity function) when calculating their results.
In this article, we propose an approach in which this proximity function takes into account not only the distances between features but also the relationships between data points.

The contribution of this paper is the following:
\begin{enumerate}
	\item Problems of regression and treatment effect estimation in the presence of externally added relationships are formulated
	\item Two methods are proposed: a modified Nadaraya-Watson kernel regression and an attention-based architecture with custom relationship-aware attention.
	Both methods are applicable to both problems.
	\item Several models based on the modified Nadaraya-Watson regression are developed and implemented: regression with relationships, regression with relationships and trainable feature weights,
	regression with relationships and features embeddings implemented using a multi-layer perceptron. Also, a transformer-like model (which we call TabRel) based on a modified attention mechanism is designed and implemented.
	\item Our models are benchmarked on synthetic and real datasets for regression, as well as on a semi-synthetic dataset for treatment effect estimation.
	The results show that while the attention-based model does not excel on most datasets for regression analysis, it performs well at treatment effect estimation.
	Nadaraya-Watson regression with learnable feature weights and relationships shows good results in most cases, outperforming the baseline model (LightGBM with relationships added to the feature set).
	Our models and scripts for experiments are available at \href{https://github.com/zuevval/tabrel}{github.com/zuevval/tabrel}
\end{enumerate}

The remainder of the paper is organized as follows: 
Section \ref{section:relatedWork} briefly reviews related research,
Section \ref{section:background} provides the necessary background,
Sections \ref{section:methods} and \ref{section:experiments} describe our proposed models and experiments, respectively,
and finally, Sections \ref{section:discussion} discusses the results and their implications, while Section \ref{section:conclusions} summarizes the contributions and suggests future work.

\section{Related Work} \label{section:relatedWork}

\subsection{Nadaraya-Watson Kernel Regression}

The Nadaraya-Watson semi-parametric smoothing estimator has been proposed by Nadaraya \cite{nadaraya_estimating_1964} and Watson \cite{watson_smooth_1964} for time series interpolation.
One of the first regression methods, it has been successfully applied in medicine \cite{nazer_nadaraya-watson_2024, elsamadony_medical_2024}, industry \cite{siraskar_application_2024}, and other domains.
The idea behind the method is to approximate the target variable $ y $ with a weighted sum of other $y$s present in the training data set, where each weight is proportional to the degree of similarity of the feature vectors.
The similarity measure is defined by a kernel function $ K $.
Common choices include Hilbert kernel \cite{devroye_hilbert_1998}, as well as standard kernels such as Epanechnikov \cite{han_nadaraya-watson_2024}, uniform, square, and Gaussian \cite{bartlett_deep_2021}; in our work, we use the Gaussian kernel (or, rather, a modified version to take relationships into account).

Besides the use of Nadaraya-Watson regression as such, various classical and deep learning methods have been developed that rely on Nadaraya-Watson regression, for example, Park et al. utilized it with boosting
\cite{park_l_2_2009}.
Recently, a regression with trainable kernels was applied to treatment effect estimation
 \cite{konstantinov_heterogeneous_2023}.

\subsection{Tabular Deep Learning}
A vast body of machine learning research is devoted to regression and classification on tabular data.
Among traditional ML methods, gradient boosting decision
trees (GBDTs) show the best performance on many benchmarks, including medical
datasets \cite{yildiz_gradient_2025}.
Convolutional neural networks, despite their ability to effectively capture feature dependencies, show modest performance on tabular data due to the lack of locality in feature ordering \cite{shwartz-ziv_tabular_2022}.
In recent years, attention-based models have been extensively
applied to tabular data analysis, including TabNet and TabTransformer \cite{somvanshi_surveyTabularDl_2024}. 
It should be noted that, while performing better than CNNs, transformers still do not
show a decisive superiority over GBDTs \cite{grinsztajn_why_2022}.
Some of the deep learning models are able to handle intersample relationships.
In 2021, Somepalli et al. introduced the SAINT (Self-Attention and INtersample attention Transformer) architecture \cite{somepalli_saint_2021}
which performs competitively with gradient boosting methods.
However, few studies focus on external, graph-like intersample
relations.
Graph Neural Networks (GNNs) \cite{khemani_reviewGnn_2024},
including Graph Attention Networks \cite{vrahatis_graphAttn_2024},
are capable of node classification (or regression).
During the inference, for any given
node $x$ those models take only connected nodes into consideration.
A graph formed by ancestral relations over patients of some clinic is typically sparse, which makes GNNs difficult to apply in this domain.

\subsection{Treatment Effect Estimation}

The individual treatment effect (ITE) with respect to a given outcome, given known patient data, is the difference between the values of that outcome when the patient received treatment and when the patient did not receive treatment.
Estimating the treatment effect given known characteristics (such as gender, age, etc.) is, from a mathematical perspective, a special case of a problem that can arise, for example, in economics when forecasting an increase in a company's revenue if certain measures are taken.
The treatment effect studies and uplift modeling in economics served as two primary drivers for the development of relevant mathematical methods \cite{zhang_unified_2021}.

The two most popular frameworks for assessing the treatment effect are T-learner (two-learner -- trains two different models to estimate the outcome) and S-learner (single learner -- trains a single model to estimate the treatment effect).
In 2019, Künzel et al. developed the X-learner \cite{kunzel_metalearners_2019} which uses 4 regression models, and in 2020, Nie et al. \cite{nie_quasi-oracle_2020} introduced the R-learner framework.
Each framework allows to use arbitrary machine learning methods for regression, such as neural networks or gradient boosting decision trees, as well as various metrics.
Knaus et al. gave a comprehensive review of the approaches and methodologies \cite{knaus_machineLearning_review_2021}.

\section{Background} \label{section:background}

\subsection{Nadaraya-Watson Kernel Regression}

Consider a sample $ \{ (x_i, y_i) \}_{i=1}^N, x_i \in \mathds R ^d, y_i \in \mathds R $.
Then, a Nadaraya-Watson Kernel regression estimator \cite{nadaraya_estimating_1964, watson_smooth_1964} $ f_{\sigma}: \mathds R^d \to \mathds R $ with parameter $ \sigma $ and exponential kernel takes the following form:

\begin{align}
	y(x) &= \sum_{i=1}^N y_i \alpha(x, x_i) \label{eq:nw} \\ 
	\alpha(x, x_i) &= \frac{K(x, x_i)}{\sum_{j=1}^N K(x, x_j)} \label{eq:nw_weights} \\
	K(x, x_i) &= \exp \left( - \frac{|| x-x_i || ^2}{ \sigma } \right) \label{eq:nw_gauss_kernel}
\end{align}

Here, $ \alpha(x, x_i) $ is the normalized distance between the input vector and the feature vector $ x_i $ of the background element $ i $.

\subsection{Treatment Effect Estimation}

Mathematically the problem of ITE estimation is formulated as the \textit{conditional average treatment effect} (CATE) estimation \cite{kunzel_metalearners_2019}.
In its basic form, we assume a distribution $ \mathcal P : (Y_i(0), Y_i(1), X_i, W_i) \sim \mathcal P $, where $i$ is the patient's index, $W_i \in \{0;1\}$ is an indicator of treatment (1) or control (0) group, $X_i \in \mathds R^d$ is a $d$-dimensional feature vector, and $Y_i(\cdot) \in \mathds R$ is the outcome, which corresponds to the patient's health status when he is treated or untreated, respectively.
CATE is defined as the difference between two outcomes:

\begin{equation}\label{eq:cateBasic}
	\tau_i \equiv \text{CATE}(X_i) \defeq Y(1) - Y(0)
\end{equation}

The term CATE is generally preferred over the term ITE to highlight that features do not explain the variability in outcomes completely, therefore, we estimate the average expected effect $ \tau_i $ conditional on the given feature set $ X_i $ rather than the expected effect for the individual.

Note that either $Y(1)$ or $Y(0)$ is not observed.
Therefore, given a data set $ \{ X_i, W_i, Y_i(W_i) \} _{i=1}^N $, a common approach is to train two estimators $ f_{\theta 1} $ and $ f_{\theta 2} $ using treated and non-treated subsets ($ f_{\theta i}: \mathds{R}^d \to \mathds R $; often parameter sets $ \theta_1 $ and $ \theta_2 $ overlap, e.g. estimators can be two networks with shared layers and different heads), or to add a binary feature  -- an indicator of treated / non-treated subgroup, and then learn a single estimator $ f_{\theta} $ for $ Y_i(W_i) $.
These methods are referred to as T-learner and S-learner, respectively \cite{kunzel_metalearners_2019}.
There are other estimators; some impute
$ Y_i(\overline W_i) $\footnote{ here $ \overline W_i \defeq \left( 0 \text{ if } W_i = 1, \quad 1 \text{ if } W_i = 0 \right) $ } assuming smoothness of $ Y(\cdot) $, and then train a second-stage parametric estimator  to either compute $ \tau_i $ directly or first estimate $ Y_i(\cdot)$ and then derive $ \tau_i $.

\section{Methods} \label{section:methods}
\subsection{Relationship-Aware Nadaraya-Watson Kernel Regression}
\subsubsection{Motivation}

One approach to building tabular transformers is to calculate attention between
rows, i.e. if the data consists of $n_{train} + n_{test}$ points, each vector (or, rather, its
embedding) "attends" each other (excluding test points that do not attend other
test points).
Response variable $y$ is concatenated with the feature vector $x$; for test
points, $y$ is usually set to $0$ to prevent data leakage while matching the training
vectors length.
Thus, the training set forms a "background" which defines the
output for each test point.
Consider the attention operation in its classic form, as introduced by Bahdanau et
al. \cite{bahdanau_neural_2016}.
Attending a data point $ \{x,y\} $ to the background set $\{x_i
^{back}, y_i^{back}\}$ can be thought of as taking a weighted sum of the background vectors, with each weight representing
the degree of similarity between $\{x, y\}$ and the background point, normalized by the sum of similarity measures between $\{ x, y \}$ and all the other background points.
Nadaraya-Watson kernel regression, discussed in
detail below in Section \ref{section:methods:relNW:formulation},
is a weighted sum of $y_i^{back}$  with each weight defined by a similarity measure (kernel) which is normalized by the sum of similarity measures between $x$ and other background feature vectors $ x_i^{back} $.
Therefore, it can be interpreted as a simplified, degenerate version of a single-layer attention module.
Some authors even call Nadaraya-Watson regression "attention"\cite{siraskar_application_2024}.

We propose a generalization of this expression to include not only the information about the similarity determined by features, but also the part defined by the relationships between samples.

\subsubsection{Formulation}\label{section:methods:relNW:formulation}

We propose to introduce a latent representation of one's genotype (or another unknown set of properties associated with some known proximity measure between data points) $ L_i \in \mathcal L $, where $ \mathcal L $ is a set of arbitrary nature.
Coupled with some similarity measure $ d: \mathcal L \times \mathcal L \to \mathds R_+ $, it can be implicitly added to the dataset if we know $ R_{ij} \defeq d(L_i, L_j) $ even when $ L_i $ is unknown or too complex to be used "as is".
For example, $ \{ R_{ij} \}_{i,j=1}^N $ may be a kinship matrix; or, if we know the patients' genotypes, we may use genome sequences as $ L_i $ and pairwise alignment scores as $ g(L_i, L_j) $.
The main challenge is to replace the estimator $ f_\theta : \mathds R^d \to \mathds R $ with another $ \tilde f_ {\tilde \theta} : \mathds R^d \times \mathds R_+^{ N \times N} \to \mathds R $.

Thus, there is an observable graph of relationships (possibly weighted; most likely sparse) that is represented by the adjacency matrix $ R_{N \times N} $ which may reflect family ties between individuals, geographic connectivity between states, and so on; in its simplest case $ R_{ij} \in \{ 0; 1 \} $, e.g. two states are connected (share a common border) or not.
One way to use relationships in the regression model is to add lines of the adjacency matrix as features, i.e. a feature vector $ x_i $ is expanded to $ \begin{pmatrix}
	x_i &  r_i
\end{pmatrix}^T $.
However, as we show in simulations in Section \ref{section:methods:nw:alternative}, this naive method works well only if matrix $ R $ reflects the underlying connections perfectly, i.e., there are no noisy or missing values in $ R $.

From our point of view, the most natural way to account for relationships is to add  $ r_{is} $ (the elements of $ R $) inside the kernel \eqref{eq:nw_gauss_kernel}  as a second term in the exponent argument, in addition to the norm of the difference between $ x_i $ and $ x_s $.
This makes the weight of $ y_i $ depend only on the relationship $ r_{is} $, and not on the $ 2N $ values from the matrix $ R $, many of which may be irrelevant. 
In section \ref{section:methods:nw:alternative} we demonstrate the robustness of this approach to noise in $ R $.

If we think of $ r_{is} $ as of similarity between two latent genotypes $ L_i $, $ L_s $ as described in Section \ref{section:intro}, then we can think of a weighted sum $ \gamma r_{is}  - \frac{1}{\sigma} || x_i - x_s ||^2 $ as a dissimilarity between two points in a unified space of latent and non-latent features $ \mathds{R}^d \times \mathcal L $.
Here, $ \gamma $ is a learnable parameter.

Therefore, \eqref{eq:nw}-\eqref{eq:nw_gauss_kernel} transforms into the following:

\begin{align}
	y(x_s) &= \sum_{i=1}^N y_i \alpha(x_s, x_i, r_{is}) \label{eq:nw_rel} \\ 
	\alpha(x_s, x_i) &= \frac{K(x_s, x_i, r_{si})}{\sum_{j=1}^N K(x_s, x_j, r_{sj})} \label{eq:nw_rel_weights} \\
	K(x_s, x_i, r_{si}) &= \exp \left( - \frac{|| x_s-x_i  || ^2}{ \sigma } + \gamma r_{si} \right) \label{eq:nw_rel_gauss_kernel}
\end{align}

Equations \eqref{eq:nw_rel} -- \eqref{eq:nw_rel_gauss_kernel} can be combined into a single expression as follows:
\begin{equation}\label{eq:nw_rel_expanded}
	y(x_s) = \frac{\sum_{i=1}^N y_i \exp \left(- \frac{1}{\sigma} || x_s - x_i ||^2 + \gamma r_{si}\right)}{\sum_{j=1}^N \exp \left(- \frac{1}{\sigma} || x_s - x_j ||^2 + \gamma r_{sj}\right)}
\end{equation}

Thus, we assume that for each new sample $ x_s $ we somehow know $ r_{si} $ for all $ i \in \overline{1, N} $.
In the limiting case where $ R $ is a matrix of all zeros (which might happen if $ R $ is defined by family ties between patients), our method reduces to standard kernel regression.

Given a training data set of $ M $ samples with known relationships, we may divide it into \textit{background} set of $ N < M $ samples and \textit{trial} set of $ M - N $ samples, then fit the model parameters by minimizing an appropriate loss function (e.g. MSE -- mean squared error) over the trial set.

\subsubsection{Toy experiment}\label{section:methods:nw:toy_experiment}

For illustration purposes, we have applied our relationship-aware kernel regression to a synthetic dataset with one-dimensional input generated as follows.
$ {x_i} $ are uniformly distributed in the interval $ [-1; 1] $; $ y_i = x_i^2 + \frac{c_i}{2} $, where $ c_i \in \{ 0; 1; 2 \} $ is a randomly assigned cluster.
We build the relationship matrix in the following way:
\begin{equation}\label{eq:r_from_clusters}
r_{ij} = \begin{cases}
	1, \quad c_i = c_j \\
	0, \quad c_i \ne c_j 
\end{cases}
\end{equation}

\begin{figure}[ht]
	\centering
	\begin{subfigure}{.5\textwidth}
		\centering
		\includegraphics[width=\textwidth]{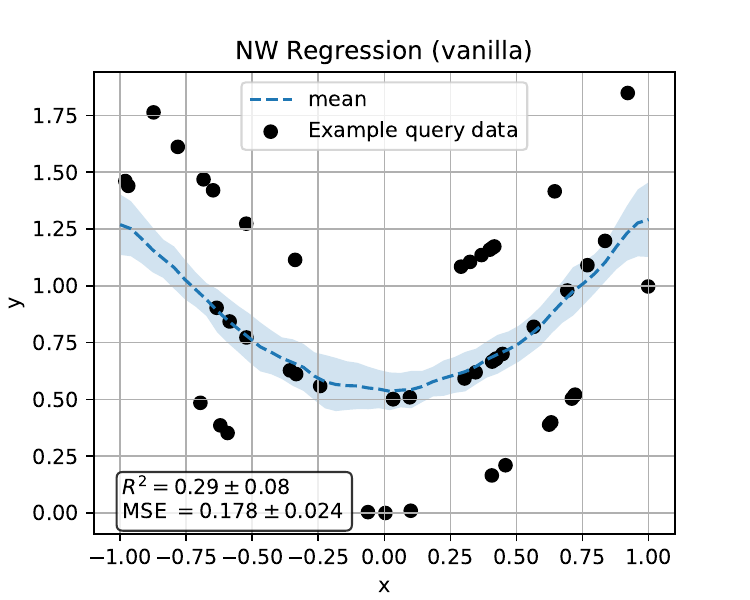}
		\caption{Vanilla}
		\label{fig:nwToyNoRel}
	\end{subfigure}%
	\begin{subfigure}{.5\textwidth}
		\centering
		\includegraphics[width=\textwidth]{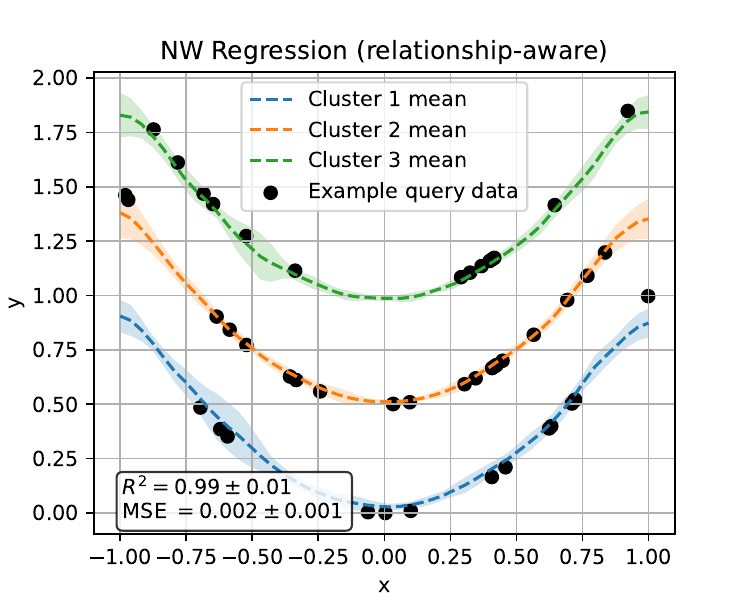}
		\caption{Relationship-aware}
		\label{fig:nwToyRel}
	\end{subfigure}
	\caption{Nadaraya-Watson regression on a toy dataset - quadratic dependency + hidden categorical additive term encoded in the relationship matrix (each element equals to 1 for two points of the same category, 0 for different categories). Confidence bands are calculated for 30 independently generated datasets}
	\label{fig:nwToyExample}
\end{figure}

Then we fit estimators \eqref{eq:nw} (Fig. \ref{fig:nwToyNoRel}) and \eqref{eq:nw_rel} (Fig. \ref{fig:nwToyRel}) using 100 data points as background and 100 as trial.
Relationship-aware estimator fits the data well ($ R^2 \approx 0.99 $ on a heldout validation set), while standard Nadaraya-Watson estimator fails as expected ($ R^2 \approx 0.29 $).

\subsubsection{Why not simply adding relationships to features}\label{section:methods:nw:alternative}

A naive way to take relationships into account mentioned in \ref{section:methods:relNW:formulation} is to concatenate $ R $ with features, i.e. if $ D $ is the initial number of features, $ X_{M \times D} $ becomes $ [X_{M \times D}, R_{M \times M} ] $.
For Nadaraya-Watson regressor, it leads to the following expression:
\begin{equation}\label{eq:nw_rel_feats}
 	y(x_s) = \frac{\sum_{i=1}^N y_i \exp \left(- \frac{1}{\sigma} \left(|| x_s - x_i ||^2 + \sum_{p=1}^{N} (r_{sp} - r_{ip})^2 \right) \right)}{\sum_{j=1}^N \exp \left(- \frac{1}{\sigma} \left(|| x_s - x_j ||^2 + \sum_{p=1}^{N} (r_{sp} - r_{jp})^2 \right) \right)}
\end{equation}

We shall denote $ \Omega_{si} \defeq \sum_{p=1}^N (r_{sp} - r_{ip})^2 $.
If our synthetic data is generated according to \eqref{eq:r_from_clusters}, $ \Omega_{si} = 0 $ when $ c_i = c_j $ and $ \Omega_{si} $ is large when $ c_i \ne c_j $, which increases weights for $ y_i $ if $ i $th and $j$th examples correspond to the same class.
Moreover, $ \Omega_{si} \overunderset{N \to \infty}{c_i \ne c_j}{\longrightarrow} \infty $.
Therefore, in this case the naive regressor \eqref{eq:nw_rel_expanded} performs not worse or even better than \eqref{eq:nw_rel_feats}.

However, consider a slightly modified version of \eqref{eq:r_from_clusters} with $R$ defined as follows
\[
r_{ij} = \begin{cases}
	1 \cdot p_{ij}, \quad c_i = c_j \\
	0, \quad c_i \ne c_j 
\end{cases}
\]
where $ p_{ij} = p_{ji} $ is a random variable which takes values $ \{0, 1\} $ with some fixed probability (we set it equal to $0.5$).
Now the estimator \eqref{eq:nw_rel_feats} will fail because $ \Omega_{si} $ can take random values, arbitrarily large when $ N \to \infty $.
Conversely, \eqref{eq:nw_rel_expanded} will still converge to the true dependency (this is confirmed empirically, see Table \ref{table:synthetic_randomR_performance}).

\subsubsection{Learnable norm}

To better handle complex dependencies, we can generalize \eqref{eq:nw_rel_expanded} by replacing the squared $ L_2 $ norm $ || x_s - x_i || ^2 $ with a quadratic form $ Q(x_s, x_i) = x_s^T W x_i $, where $ W_{D\times D} $ is learnable.
In its simplest form, $ W = \mathbb I w^2 $, where $  w $ is a learnable vector and $ \mathbb I $ denotes an identity matrix.
$ w $ is squared to keep $ Q $ positive semidefinite.
This norm can help to discard noisy components of $ x $ and balance.
Potentially we could make off-diagonal values of $ W $ learnable, too, but it would have required complex procedures to ensure $ Q $ is positive semidefinite.

\subsubsection{ Regression on feature vector embeddings }\label{section:methods:nw:mlp}

Regression \eqref{eq:nw_rel_expanded} has only 2 learnable parameters.
If the norm is learnable, the number of parameters increases to $ 2 + D $, which, however, is still small, especially for low-dimensional feature spaces.
In some cases it is an advantage, but it impacts the generalization ability of the model.
We attempted to increase the model's power by replacing the distance between feature vectors with the distance between feature vector embeddings obtained using a three-layer perceptron with the ReLU activation function.
The perceptron parameters are learnable.

\subsection{TabRel: Relationship-Aware Tabular Transformer}
\subsubsection{Model Architecture}\label{section:method:tabrel:architecture}
We developed a new transformer-like architecture for regression on tabular data with relationships which we call TabRel (\textbf{Tab}ular Transformer with \textbf{Rel}ationships).

In the core of TabRel is a stack of relationship-aware transformer encoder layers, each layer applies attention between data points.
A common choice for graphs is GATv2 layer \cite{brody_howAttentiveAreGATs_2022}, however, it only allows passing "messages" between connected vertices, and our relationship graph may be sparse and contain multiple connectivity components (e.g., typically only a few patients in a hospital are relatives).
Therefore, we have constructed a custom Relational Multi-Head Attention (RMHA) layer (described in Appendix \ref{appendix:implementationDetails}), allowing every point $ i $ in the training set to attend to any other point $ j $ even if $ r_{ij}=0 $.

We used RTDL NumEmbeddings \cite{gorishniy_embeddings_2023} to project input features into a higher dimensional space.
After the initial embedding and a linear input projection, we have put
a stack of RMHA layers and a linear output layer.

As in original Transformer, we added dropout \cite{hinton_dropout_2012} layers before each attention layer for normalization.

Although TabRel is a much more flexible model than the relationship-aware Nadaraya-Watson regression estimator, in our experiments it almost always performs similarly or worse as shown in Section \ref{section:experiments}.
Moreover, the current design requires the test set and its relationships with the training samples be known in advance, prior to training.
Therefore, TabRel has a potential for improvement, and modifying its architecture is a promising direction for future studies.



\section{Experiments}\label{section:experiments}

\subsection{Synthetic regression datasets}\label{section:experiments:synthetic}

We conducted experiments on two toy datasets, one of them mentioned in \ref{section:methods:nw:toy_experiment}.
Each time we sampled 300 points (200 for the train set, 100 for the test set).
The single feature $ x $ values were drawn from a uniform distribution on $ [-1; 1] $.
The response variable values were constructed as follows.

\begin{enumerate}
	\item Parabolas: as described in Section \ref{section:methods:nw:toy_experiment}, the cluster index $ i_c $ has been added to a quadratic function $ y_{base} = x^2 $
	\item Step functions:  instead of quadratic dependency, we use the signum $ y_{base} = sign(x) $. Then, similarly to parabolas, we added the cluster index.
\end{enumerate}

\begin{table}[h!]
	\centering
	\renewcommand{\arraystretch}{1.3}
	\begin{tabular}{|l|cc|cc|cc|cc|}
		\hline
		\multirow{3}{*}{\textbf{Method}} 
		& \multicolumn{4}{c|}{\textbf{Parabolas}} 
		& \multicolumn{4}{c|}{\textbf{Step functions}} \\ 
		\cline{2-9}
		& \multicolumn{2}{c|}{MSE} & \multicolumn{2}{c|}{$R^2$} 
		& \multicolumn{2}{c|}{MSE} & \multicolumn{2}{c|}{$R^2$} \\
		\cline{2-9}
		& mean & std & mean & std 
		& mean & std & mean & std \\
		\hline\hline
		NW, no rel-s info &
		0.178 & 0.024 & 0.29 & 0.08 &
		0.26 & 0.08 & 0.78 & 0.06 \\
		NW, rel-s features &
		0.093 & 0.012 & 0.63 & 0.05 &
		0.97 & 0.04 & 0.15 &  0.05 \\
		NW $ + $ relations &
		\textbf{0.002} & 0.001 & \textbf{0.99} & 0.01 &
		0.12 & 0.04 & 0.89 & 0.04 \\
		LightGBM, no rel-s info &
		0.183 & 0.015 & 0.27 & 0.06&
		0.20 & 0.03 & 0.82 & 0.03  \\
		LightGBM, rel-s features &
		0.012 & 0.004 & 0.93  & 0.02  &
		0.05 & 0.03 & 0.96 & 0.03  \\
		TabRel  &
		0.003 & 0.002 & 0.99 & 0.01 &
		\textbf{0.02} & 0.02 & \textbf{0.98} & 0.02 \\  
		\hline
	\end{tabular}
	\caption{Performance (mean and standard deviation of MSE and $R^2$) on synthetic datasets with one-dimensional features deterministic $R$ matrix.
	The best result in each column is highlighted in \textbf{bold}. 
	Here, "parabolas" and "step functions" are dependencies described in \ref{section:experiments:synthetic}: $ y_i = x^2 + c_i $ and $ y_i = sign(x_i) + c_i $, respectively }
	\label{table:synthetic_performance}
\end{table}

In Table \ref{table:synthetic_performance}, we display the results of experiments with synthetic datasets where target variable is highly dependent on the structure of the relationship matrix $R$ defined by equation \eqref{eq:r_from_clusters}.
Table \ref{table:synthetic_randomR_performance} shows similar results for $R$ with randomness introduced as described in \ref{section:methods:nw:alternative}.

We denote the compared methods in tables as follows:
\begin{enumerate}[noitemsep]
	\item \textbf{NW, no rel-s info}: Nadaraya-Watson kernel regression without relationships taken into account (a single learnable parameter $ \sigma $)
	\item \textbf{NW + relations}: Nadaraya-Watson regression with relationships added as in Equation \eqref{eq:nw_rel_expanded}
	\item \textbf{NW, rel-s features}: Nadaraya-Watson regression with $ R $ concatenated with features as in Equation \eqref{eq:nw_rel_feats} 
	\item \textbf{NW + relations, MLP embeddings}: Nadaraya-Watson regression on embeddings obtained with a multi-layer perceptron as described in \ref{section:methods:nw:mlp}.

	Adding a perceptron to the model allows selecting hyperparameters (hidden layer size, output layer size, dropout ratio); we tuned hyperparameters using the Optuna framework \cite{akiba_optuna_2019}.
\end{enumerate}

\begin{table}[h!]
	\centering
	\renewcommand{\arraystretch}{1.3}
	\begin{tabular}{|l|cc|cc|cc|cc|}
		\hline
		\multirow{3}{*}{\textbf{Method}} 
		& \multicolumn{4}{c|}{\textbf{Parabolas}} 
		& \multicolumn{4}{c|}{\textbf{Step functions}} \\
		\cline{2-9}
		& \multicolumn{2}{c|}{MSE} & \multicolumn{2}{c|}{$R^2$} 
		& \multicolumn{2}{c|}{MSE} & \multicolumn{2}{c|}{$R^2$} \\
		\cline{2-9}
		& mean & std & mean & std 
		& mean & std & mean & std \\
		\hline\hline
		NW, no rel-s info & 0.172 & 0.012 & 0.32 & 0.05 & 0.187 & 0.022 & 0.57 & 0.26 \\
		NW, rel-s features & 0.946 & 0.068 & -2.73 & 0.40 & 1.169 & 0.246 & -1.48 & 1.30 \\
		NW + relations & 0.020 & 0.005 & 0.92 & 0.02 & 0.039 & 0.024 & 0.94 & 0.02 \\
		LightGBM, no rel-s info & 0.182 & 0.013 & 0.28 & 0.06 & 0.180 & 0.013 & 0.56 & 0.29 \\
		LightGBM, rel-s features & \textbf{0.019} & 0.003 & \textbf{0.92} & 0.01 & \textbf{0.019} & 0.003 & \textbf{0.95} & 0.03 \\
		TabRel & 0.99 & 0.24 & -2.89 & 0.96 & 1.2 & 0.3 & -1.57 & 1.50 \\
		\hline
	\end{tabular}
	\caption{Performance (mean and standard deviation of MSE and $R^2$) on synthetic datasets with one-dimensional features and randomness in $ R $ matrix.
	The best result in each column is highlighted in \textbf{bold}
	}
	\label{table:synthetic_randomR_performance}
\end{table}

Additionally, we conducted similar experiments on synthetic datasets with two features and deterministic $ R $.

\begin{table}[h!]
	\centering
	\renewcommand{\arraystretch}{1.3}
	\begin{tabular}{|ll|cc|cc|cc|}
		\hline
		\multicolumn{2}{|l|}{\textbf{Method}} 
		& \multicolumn{2}{c|}{\textbf{Linear}} 
		& \multicolumn{2}{c|}{\textbf{Square}} 
		& \multicolumn{2}{c|}{\textbf{Sin}} \\
		\cline{3-8}
		\multicolumn{2}{|l|}{} 
		& mean & std 
		& mean & std 
		& mean & std \\
		\hline\hline
		NW, no rel-s info, & n=300 & 0.8861 & 0.0176 & 0.6180 & 0.0596 & 0.7349 & 0.0454 \\
		$ L_2 $ norm & n=1000 & 0.9024 & 0.0064 & 0.6594 & 0.0297 & 0.7729 & 0.0161 \\
		\hline
		NW, no rel-s info, & n=300 
		& 0.8902 & 0.0176 
		& 0.6365 & 0.0551 
		& 0.7474 & 0.0441 \\
		learnable norm & n=1000 
		& 0.9025 & 0.0063 
		& 0.6623 & 0.0289 
		& 0.7747 & 0.0160 \\
		\hline
		\multirow{2}{*}{NW, rel-s features} & n=300 & -0.1619 & 0.0868 & -0.926 & 0.2373 & 0.3754 & 0.1459 \\
		& n=1000 & -0.16 & 0.0421 & -0.902 & 0.1244 & -0.3593 & 0.074 \\
		\hline
		NW $ + $ relations, & n=300 
		& 0.9839 & 0.0038 
		& 0.9695 & 0.0061 
		& 0.9777 & 0.0045 \\
		$ L_2 $ norm & n=1000 
		& 0.9935 & 0.0011 
		& 0.9874 & 0.0013 
		& 0.9910 & 0.0011 \\
		\hline
		NW $ + $ relations, & n=300 
		& 0.9701 & 0.0069 
		& 0.9349 & 0.0160 
		& 0.9532 & 0.0116 \\
		learnable norm  & n=1000 
		& 0.9860 & 0.0014 
		& 0.9679 & 0.0028 
		& 0.9774 & 0.0019 \\
		\hline
		NW $ + $ relations, & n=300 
		& 0.9806 & 0.0051 
		& 0.9498 & 0.0097 
		& 0.9692 & 0.0093 \\
		MLP embeddings  & n=1000 
		& 0.9901 & 0.0012 
		& 0.9679 & 0.0022 
		& 0.9822 & 0.0025 \\
		\hline
		LightGBM, & n=300 & 0.8806 & 0.0220 & 0.6003 & 0.0702 & 0.7233 & 0.0537 \\
		no rel-s info & n=1000 & 0.8871 & 0.0099 & 0.6056 & 0.0392 & 0.7361 & 0.0219 \\
		\hline
		LightGBM, & n=300 & \textbf{0.9859} & 0.0035 & \textbf{0.9880} & 0.0024 & \textbf{0.9835} & 0.0025 \\
		rel-s features & n=1000 & \textbf{0.9957} & 0.0006 & \textbf{0.9965} & 0.0005 & \textbf{0.9954} & 0.0008 \\
		\hline
		\multirow{2}{*}{TabRel} & n=300 & 0.8668 & 0.2762 & 0.8597 & 0.1396 & 0.9327 & 0.0571 \\
		& n=1000 & 0.9486 & 0.1042 & 0.9279 & 0.0311 & 0.9392 & 0.1388 \\
		\hline
	\end{tabular}
	\caption{Results ($ R^2 $) for synthetic data with two features.
	The best result in each column is highlighted in \textbf{bold} for $ n=300 $ and $ n=1000 $ separately
	}
	\label{tab:results_synthetic2d}
\end{table}

We run each algorithm on datasets of size $ n=300 $ and $ n=1000 $ independently.
Background, trial and validation sets each constitute approximately $\frac{1}{3}$ of the total dataset size.
Table \ref{tab:results_synthetic2d} contains the results.
In Table \ref{tab:results_synthetic2d}, dependencies are the following:
\begin{itemize}
	\item Linear: $ y_i = x_{1i} + 2 x_{2i} + 0.5 c_i $
	\item Square: $ y_i = x_{1i} + 0.5 x_{2i}^2 + 0.5 c_i $
	\item Sin: $ y_i = x_{1i} + \sin\left( x_{2i}\right) + 0.5 c_i $
\end{itemize}

where $ c_i $ is the cluster label (0, 1, or 2) of $i$th data point.

To test the ability of kernel regression with learnable norm to handle noisy inputs, we have conducted experiments with multidimensional synthetic data: 
\begin{equation}\label{eq:synthetic7dData}
	x_i \in \mathds R ^7, x_{ij} \sim U[-1;1], y_i = \cos \left( x_{i0} \right) + \cos\left( x_{i1} \right) + x_{i3} + c_i
\end{equation}
Therefore, in this synthetic dataset with multidimensional features some are noisy (not correlated with the target variable).

Table \ref{table:synthetic_noisy_7d} contains the results for this dataset.

\begin{table}[h!]
	\centering
	\renewcommand{\arraystretch}{1.3}
	\begin{tabular}{|l|cc|cc|}
		\hline
		\multirow{2}{*}{\textbf{Method}}
		& \multicolumn{2}{c|}{MSE} & \multicolumn{2}{c|}{$R^2$} \\
		\cline{2-5}
		& mean & std & mean & std \\
		\hline\hline
		NW, no rel-s info & 0.85 & 0.108 & 0.31 & 0.04 \\
		NW, rel-s features & 0.61 & 0.07 & 0.53 & 0.05 \\
		NW + relations, $L_2$ norm  & 0.23 & 0.04 & 0.82 & 0.02 \\
		NW + relations, learnable norm  & \textbf{0.07} & 0.01 & \textbf{0.95} & 0.01 \\
		NW + relations, MLP embeddings & 0.0714 & 0.0002 & 0.9434 & 0.0002  \\
		LightGBM, no rel-s info & 0.85 & 0.08 & 0.33 & 0.06 \\
		LightGBM, rel-s features & 0.10 & 0.02 & 0.92 & 0.02 \\
		TabRel & 2.8 & 1.2 & -1.2 & 1.0 \\
		\hline
	\end{tabular}
	\caption{Performance (mean and standard deviation of MSE and $R^2$) on a synthetic 7-dimensional dataset with randomness in $ R $ matrix and noise in features (target variable defined in Equation \eqref{eq:synthetic7dData}).
	The best result in each column is highlighted in \textbf{bold}
	}
`\label{table:synthetic_noisy_7d}
\end{table}

\subsection{Regression on Publicly Available Data}

To our knowledge, there are no publicly available regression datasets which that simultaneously contain features, target variable values, and a graph of intersample relationships.
This paper uses several publicly available regression datasets for which relatedness graphs can be constructed using external sources.

We tested the concept of relationship-aware Nadaraya-Watson regression on the Countries Life Expectancy dataset which provides per-country, per-year statistics.
It is designed to predict life expectancy based on a range of health-related characteristics.
We defined relationships by the mutual arrangement of countries (more details are provided in the Appendix \ref{appendix:experimentDetails}).

We predicted the life expectancy both based on features + relationships and features only.
The relationship matrix is designed as follows: for two states $ i, j $

\begin{equation}\label{eq:r_countries}
	r_{ij} = \begin{cases}
		1, \quad \text{ countries } i,j \text{ share a common border } \\
		0, \quad \text{ countries } i,j \text{ do not share a common border}
	\end{cases}
\end{equation}

For comparison, we trained standard Nadaraya-Watson regression without relationship information, LightGBM, LightGBM with relationships added to the feature set, and our transformer (TabRel).
The results are represented in the table \ref{table:regression_performance}.

In addition to the Countries Life Expectancy dataset, we used the Birds Genetic Diversity dataset which contains data records (body mass, breeding range, and genetic richness) for approximately 400 birds species.
We impose an external relatedness graph based on the taxonomy of birds.
For more details, see the Appendix \ref{appendix:experimentDetails}.

Table \ref{table:regression_performance} lists results obtained for these two datasets.

Adding the relationship matrix $ R $ slightly improves the performance for Birds and Life Expectancy;
concatenating $R$ with features improves the performance of Nadaraya-Watson Regression on Birds dataset even further, but fails on the Life Expectancy.
LightGBM always performs worse than Nadaraya-Watson Regression.

\begin{table}[h!]
	\centering
	\renewcommand{\arraystretch}{1.3}
	\begin{tabular}{|l|cc|cc|cc|cc|}
		\hline
		\multirow{2}{*}{\textbf{Method}} & 
		\multicolumn{4}{c|}{\textbf{Life Expectancy}} & 
		\multicolumn{4}{c|}{\textbf{Birds}} \\
		\cline{2-9}
		& \multicolumn{2}{c|}{MSE} & \multicolumn{2}{c|}{$R^2$} & 
		\multicolumn{2}{c|}{MSE} & \multicolumn{2}{c|}{$R^2$} \\
		\cline{2-9}
		& mean & std & mean & std & mean & std & mean & std \\
		\hline\hline

		NW, no rel-s info & 34.72 & 7.77 & 0.44 & 0.16 & 0.240 & 0.035 & 0.162 & 0.088 \\
		NW, rel-s features & 1357.2 & 92.5 & -20.5 & 2.0 & \textbf{0.226} & 0.036 & \textbf{0.214} & 0.083 \\
		NW + relations, $L_2$ norm & 26.95 & 3.85 & 0.57 & 0.08 & 0.235 & 0.033 & 0.178 & 0.096 \\
		NW + rel-s, learnable norm & 26.40 & 3.02 & 0.58 & 0.06 & 0.236 & 0.032 & 0.176 & 0.090 \\
		NW + relations + MLP & 27.76 & 3.42 & 0.56 & 0.07 & 0.237 & 0.032 & 0.172 & 0.097 \\
		LightGBM, no rel-s info & \textbf{22.11} & 1.75 & \textbf{0.65} & 0.04 & 0.257 & 0.032 & 0.102 & 0.083 \\
		LightGBM, rel-s features & 22.11 & 1.75 & 0.65 & 0.04 & 0.257 & 0.032 & 0.104 & 0.080 \\
		TabRel & 40.9 & 13.4 & 0.43 & 0.23 & 0.28 & 0.03 & 0.03 & 0.13 \\
		\hline
	\end{tabular}
	\caption{Performance (mean and standard deviation of MSE and $R^2$) on the Life Expectancy and Birds datasets.
	The best result in each column is highlighted in \textbf{bold}
	}
\label{table:regression_performance}
\end{table}

On the Life Expectancy dataset, the best method is LightGBM; however, adding relationships as features neither improves nor worsens its performance.
The best method for the Birds dataset is the Nadaraya-Watson regression with relationships added directly to the feature set.

\subsection{Relationship-Aware Treatment Effect Estimation} \label{section:experiments:TE_estimation}

So long as there are no open datasets for treatment effect estimation with known relatedness between sample, we have constructed one artificially from IHDP dataset.
IHDP contains over 700 recordings with 25 covariates, both numerical and categorical.
We have removed the covariate \texttt{x4} which is a categorical column with 4
categories, and constructed $R$ as follows: $r_{ij} = 1$ if samples are of the same category
as defined by \texttt{x4}, $0$ otherwise.

Then, we applied our regression estimators (Nadaraya-Watson with and without relationships taken into consideration, LightGBM, LightGBM with relationships $ R $ concatenated with features $ X $, and TabRel) to treatment effect estimation in three settings:
\begin{enumerate}
	\item S-learner, as described in Section \ref{section:intro}: $ \tau_S(x) \defeq \hat \mu(x, 1) - \hat \mu(x, 0)  $, where $ \hat \mu $ is a regressor trained to predict the outcome $ Y $ conditional on features $ X $ and the treatment indicator $ W_i \in \{ 0; 1 \} $
	\item T-learner, as described in Section \ref{section:intro}: $ \tau_T(x) \defeq \hat \mu_1(x) - \hat \mu_0(x) $, where $ \mu_1 $ and $ \mu_0 $ are two regressors trained separately on patients from treated and control groups, respectively.
	\item X-learner \cite{kunzel_metalearners_2019}: $ \hat \tau_X(x) = \frac{1}{2} \left(\tau_0(x) + \tau_1(x)\right) $, where $ \tau_0 $ and $ \tau_1 $ are regressors trained on the whole dataset to estimate the treatment effect; $ \hat \mu_1 $ and $ \hat \mu_2 $ are used to impute unknown outcomes.
\end{enumerate}

Table \ref{table:ihdp_performance} shows results for treatment effect estimation.
PEHE (precision of estimation of heterogenous effects) is a commonly used \cite{lee_pehe_2023} metric defined as the average squared difference between estimated and true CATE.

\begin{table}[h!]
	\centering
	\renewcommand{\arraystretch}{1.3}
	\begin{tabular}{|l|cc|cc|cc|}
		\hline
		\multirow{2}{*}{\textbf{Method}} 
		& \multicolumn{2}{c|}{\textbf{S}} & \multicolumn{2}{c|}{\textbf{T}} & \multicolumn{2}{c|}{\textbf{X}} \\
		\cline{2-7}
		& mean & std & mean & std & mean & std \\
		\hline\hline
		NW, no rel-s info & 15.3 & 2.3 & 61.2 & 3.0 & 7.0 & 1.5 \\
		NW, rel-s features & 16.9 & 2.0 & 51.9 & 3.0 & 7.1 & 1.6 \\
		NW + relations, $ L_2 $ norm & 14.8 & 2.1 & 66.8 & 3.5 & 6.9 & 2.0 \\
		NW + relations, learnable norm & 8.6 & 1.8 & 4.6 & 0.4 & 5.2 & 0.4 \\
		NW + relations + MLP & \textbf{3.9} & 0.5 & \textbf{3.3} & 0.3 & 4.1 & 0.5 \\
		LightGBM, no rel-s info & 4.4 & 0.3 & 3.6 & 0.3 & 4.0 & 0.2 \\
		LightGBM, rel-s features & 4.5 & 0.3 & 3.6 & 0.3  & \textbf{4.0} & 0.2 \\
		TabRel & 24.3 & 3.0 & 5.8 & 1.4 & 5.5 & 1.4 \\
		\hline
	\end{tabular}
	\caption{PEHE scores on the IHDP dataset.
	Column names correspond to learners names (S-learner, T-learner, X-learner).
	The best result in each column is highlighted in \textbf{bold}
	} \label{table:ihdp_performance}
\end{table}





\section{Discussion} \label{section:discussion}

One inherent limitation of Nadaraya-Watson regression \eqref{eq:nw}, which also holds for the relationship-aware modification \eqref{eq:nw_rel}, is that the output is always a convex linear combination of background outputs $ y_i $, since weights \eqref{eq:nw_weights}, \eqref{eq:nw_rel_weights} are positive.
This can be mitigated by replacing $ y_i $ with some function $ f_\theta (y_i) $ with learnable parameter $ \theta $; experimenting with different functions might be a promising future direction.
Similarly, instead of fixed $ L_2 $ or learnable norm of $ x_i - x_s $, one may use transformed input features $ || g_\eta (x_s) - g_\eta (x_i) || $, where $g $ is some function with learnable parameter s vector $ \eta $, e.g., a multi-layer perceptron, as described in \ref{section:methods:nw:mlp}.
In this paper, we have demonstrated that learnable feature weights improve the performance of Nadaraya-Watson regression in the presence of noisy features, which corresponds to $ g_\eta = \eta, \hspace{1em} \eta \in \mathds R^d $ (linear transformation).
We also experimented with embeddings obtained with 3-layer perceptron; in our experiments, its performance sometimes surpasses a simpler regression with learnable feature weights (in terms of $ R^2 $ on the validation set or PEHE for treatment effect data), but without a careful tuning of hyperparameters it is worse.

Another caveat is that in the form \eqref{eq:nw_rel_weights} relationships between background samples are ignored during training.
Both relationship-aware transformer and kernel regression in the form \eqref{eq:nw_rel_feats} take all relationship into consideration, but first is difficult to train and second has limitation discussed in \ref{section:methods:nw:alternative} (it works correctly only if $ i $th and $j$th rows in $R$ are almost the same when $ c_i = c_j $).
Future studies may include adding inter-background relationships into the Nadaraya-Watson regression model.

For now, the relationship-aware transformer, while being the most powerful and flexible model, falls behind other methods in many cases, e.g. in Table \ref{table:synthetic_randomR_performance} and Table \ref{table:synthetic_noisy_7d} the transformer shows an average value of $ R^2 $ less than zero on synthetic datasets while other methods (relationship-aware Nadaraya-Watson regression, LightGBM with relationships added to features) both have $ R^2 > 0.95 $.

Nadaraya-Watson regression with relationships added to the feature set (Equation \eqref{eq:nw_rel_feats}) in many cases shows comparable performance to the relationship-aware kernel regression (Equation \eqref{eq:nw_rel_expanded}), but, as discussed previously in \ref{section:methods:nw:alternative} (and shown in Table \ref{table:synthetic_randomR_performance}), it can easily fail if not all underlying relationships are captured in the $ R $ matrix.
Table \ref{table:regression_performance} (regression on real datasets) shows that, while being even better than \eqref{eq:nw_rel_expanded} in one case, it fails completely for the Life Expectancy dataset.

Kernel regression with a learnable norm has proven its ability to suppress noisy features, as shown on synthetic data (Table \ref{table:synthetic_noisy_7d}), but our "real" benchmark datasets in Table \ref{table:regression_performance} are low-dimensional (1-2 features), therefore, we cannot gain any advantage by introducing learnable feature weights there.

For treatment effect estimation on IHDP (Table \ref{table:ihdp_performance}), based on the estimators we have tested, there is no single best framework in terms of PEHE on validation set.
It was unexpected because X-learner was introduced as a better alternative to S- and T-learners.
However, for all methods results within X-learner framework are better or not substantially worse than within the other two frameworks; and even the authors of X-learner caution that there is no single best choice for any treatment effect estimation dataset \cite{kunzel_metalearners_2019}.

\section{Conclusion} \label{section:conclusions}

In this paper, we propose the relationship-aware Nadaraya-Watson regression and the relationship-aware tabular transformer (TabRel).
We describe their application to regression on synthetic datasets, datasets with externally added relationship graphs, and
treatment effect estimation.
Furthermore, we discuss the theoretic benefits and drawbacks of each model for different types of data, describe their implementation details, and report their performance on several benchmarks.

We applied our models to synthetic data, three real-world regression datasets (created by combining tabular data with spatial or taxonomic data), and a semi-synthetic treatment effect estimation dataset constructed from the IHDP, which is also semi-synthetic.
In many cases, our models perform better than LightGBM with relationship matrix added to the feature set or LightGBM without relationships taken into account.
However, we conclude that the tabular transformer is often unable to learn the dependency behind the dataset at a sufficient level, and therefore is not a universal solution (at least in the current form).

\section{Data Availability}

As of June 17, 2025, all datasets mentioned in this paper and its appendices are publicly available.

\begin{enumerate}
	\item Countries Life Expectancy: \href{https://www.kaggle.com/datasets/amirhosseinmirzaie/countries-life-expectancy}{kaggle.com/datasets/amirhosseinmirzaie/countries-life-expectancy}
	\item Naturalearth (countries topology): \href{https://www.naturalearthdata.com/}{naturalearthdata.com}
	\item Bird Genetic Diversity: \href{https://www.kaggle.com/datasets/mexwell/bird-genetic-diversity}{kaggle.com/datasets/mexwell/bird-genetic-diversity}
	\item Animal Analyzing (taxonomy): \href{https://www.kaggle.com/datasets/willianoliveiragibin/animal-analyzing}{kaggle.com/datasets/willianoliveiragibin/animal-analyzing}
	\item IHDP\cite{louizos_cevae_2017}: \href{https://github.com/AMLab-Amsterdam/CEVAE}{github.com/AMLab-Amsterdam/CEVAE}
\end{enumerate}

\section{Code Availability}

The code is available at \href{https://github.com/zuevval/tabrel}{github.com/zuevval/tabrel}

\printbibliography

\newpage
\include{appendix.tex}

\end{document}

%% file: appendix.tex
\appendix
\section{Experiments Details}\label{appendix:experimentDetails}
\subsection{Countries Life Expectancy Dataset}

\begin{figure}[ht]
	\centering
	\includegraphics[width=.8\textwidth]{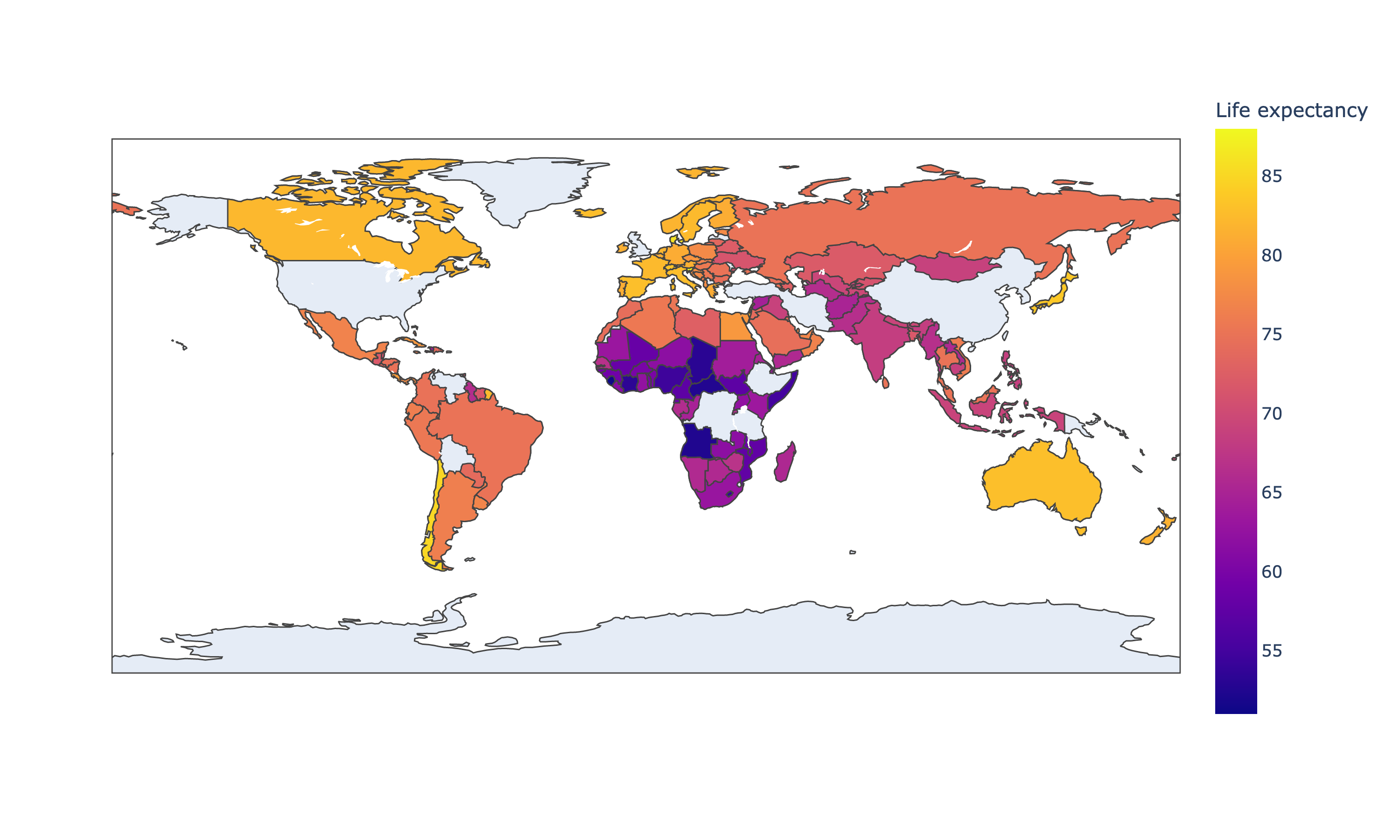}
	\caption{A choropleth for the target variable in the Life Expectancy dataset in 2015.
		Light gray shade indicates missing values }
	\label{fig:literacy_choropleth}
\end{figure}

The Countries Life Expectancy Dataset provides country-wide statistics such as
 the percentage of 1-year-olds immunized against Hepatitis B and polio, the number of deaths caused by AIDS per 1000 people and so on as well as the life expectancy from 2000 to 2015
 (Fig. \ref{fig:literacy_choropleth}).

We have selected five continuous variables -- Hepatitis B, Polio, Diphtheria, HIV/AIDS, BMI -- and fitted two estimators for the further variable conditional on the latter: relationship-aware kernel regression and conventional Nadaraya-Watson regression.
The relationship matrix is built according to the equation \eqref{eq:r_countries}.
The Naturalearth database was used to retrieve countries' topology; we considered only borders between mainlands, e.g. Brazil shares a common border with French Guiana, but we do not consider Brazil and France as neighbors.

30 countries were selected for the trial set and 30 others for the validation set; remaining constitute the background set.
We repeated the fitting multiple times using different random seeds.

As Table \ref{table:regression_performance} demonstrates, this dataset is an example where Nadaraya-Watson regression with relationship matrix rows concatenated with features is unable to approximate the dependency, whereas LightGBM and our relationship-aware regression perform acceptably.

\subsection{Bird genetic diversity}

\begin{figure}[ht]
	\centering
	\includegraphics[width=\textwidth]{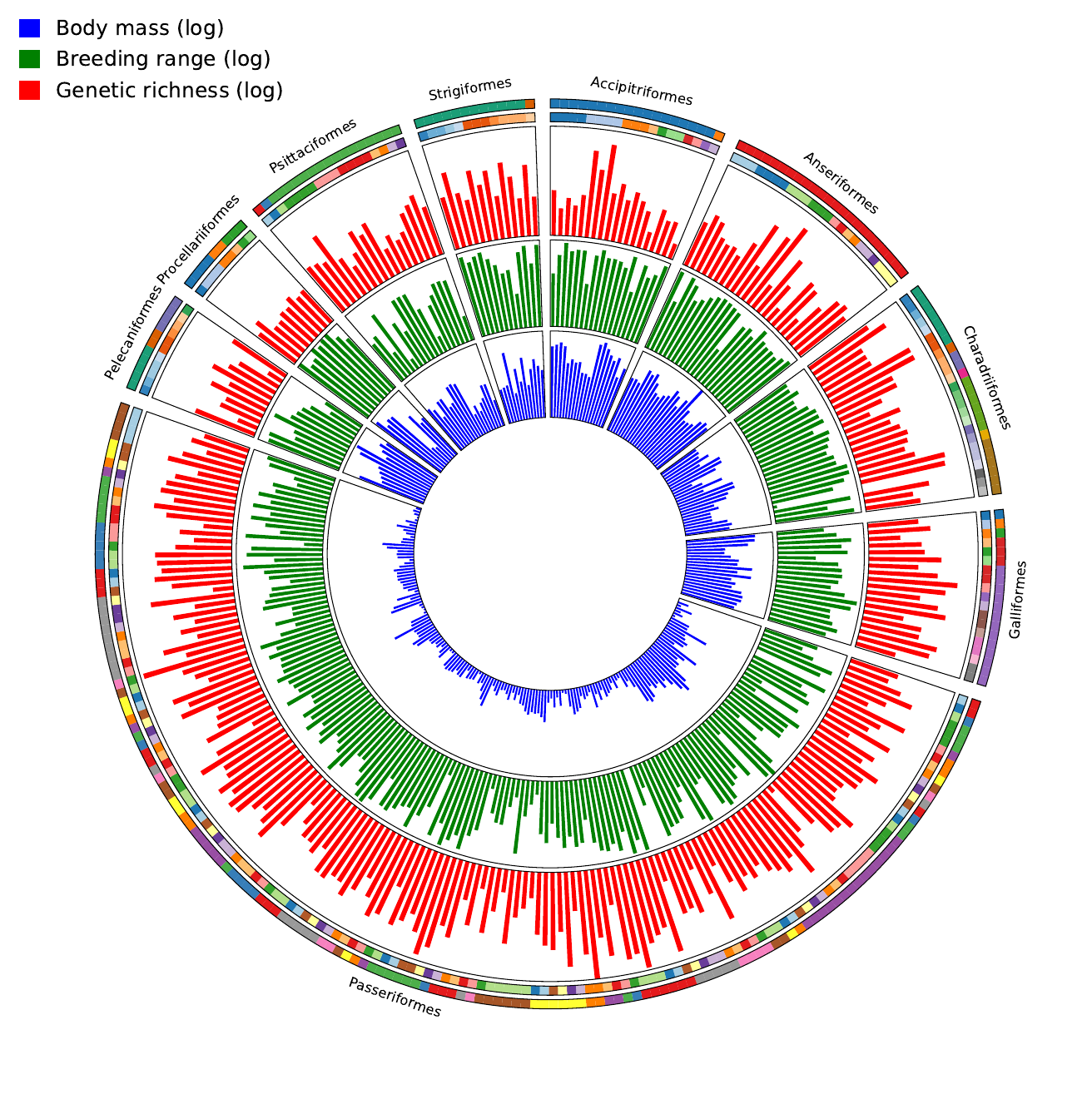}
	\caption{
		Birds genetic diversity (Birds dataset). Each sector corresponds to an order. Outermost track visualizes the family, second-to-outermost -- genus (same color within each order means the same family or genus). Orders with few species present in the dataset (<9) are not shown
	}
	\label{fig:circos_birds}
\end{figure}

We used Animals Analyzing dataset to retrieve order, family, and genus for each species in the Birds Genetic Diversity (Birds) dataset.
Figure \ref{fig:circos_birds} displays the combined data structure, where each sector is an order and different colors in two outermost tracks correspond to different families and genus, respectively (colors may repeat, but two neighboring species are of the same color only if they are of the same taxa, and they are sorted by taxa).
We used genetic richness as the target variable, and body mass along with breeding range as features (all three on a logarithmic scale).

The correlation between features and the target variable in this dataset is low, so the average $ R^2 $ on the validation subset $ \approx 0.1 - 0.2 $ for both LigthGBM and Nadaraya-Watson regression.

\subsection{Checking the informativeness of the relationship matrix}

\begin{figure}[ht]
	\centering
	\begin{subfigure}{.5\textwidth}
		\centering
		\includegraphics[width=\textwidth]{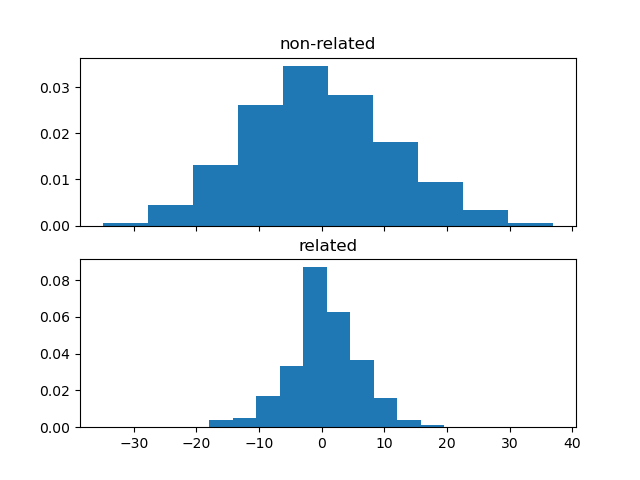}
		\caption{Life Expectancy}
		\label{fig:world_pairs_hists}
	\end{subfigure}%
	\begin{subfigure}{.5\textwidth}
		\centering
		\includegraphics[width=\textwidth]{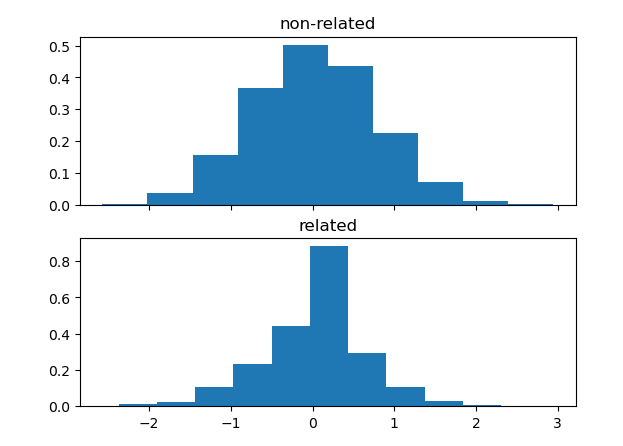}
		\caption{Birds}
		\label{fig:birds_pairs_hists}
	\end{subfigure}
	\caption{Empirical PDFs of differences between target variables in pairs with and without the relationship for Life Expectancy and Birds datasets}
	\label{fig:birds_and_lifeExpect_hists}
\end{figure}

To verify the informativeness of the relationship matrices defined for the Life Expectancy and Birds datasets, we calculated the absolute differences in the target variable $ y $ between every two points for each dataset (Life Expectancy dataset contains records for multiple years, and for this test we used data points with year=2015).
Then, we compared two empirical distributions: of all related samples (i.e., neighboring countries or birds from the same order) and of all unrelated samples (Fig. \ref{fig:birds_and_lifeExpect_hists}).
The two-sided Kolmogorov-Smirnov test yielded p-values of approximately $ 4 \times 10^{-9} $ and $ 6 \times 10^{-17} $, respectively.
This supports our assumption that relationship matrices are informative.

\section{Implementation Details}  
\label{appendix:implementationDetails}

\begin{figure}[ht]
	\centering
	\includegraphics[width=\textwidth]{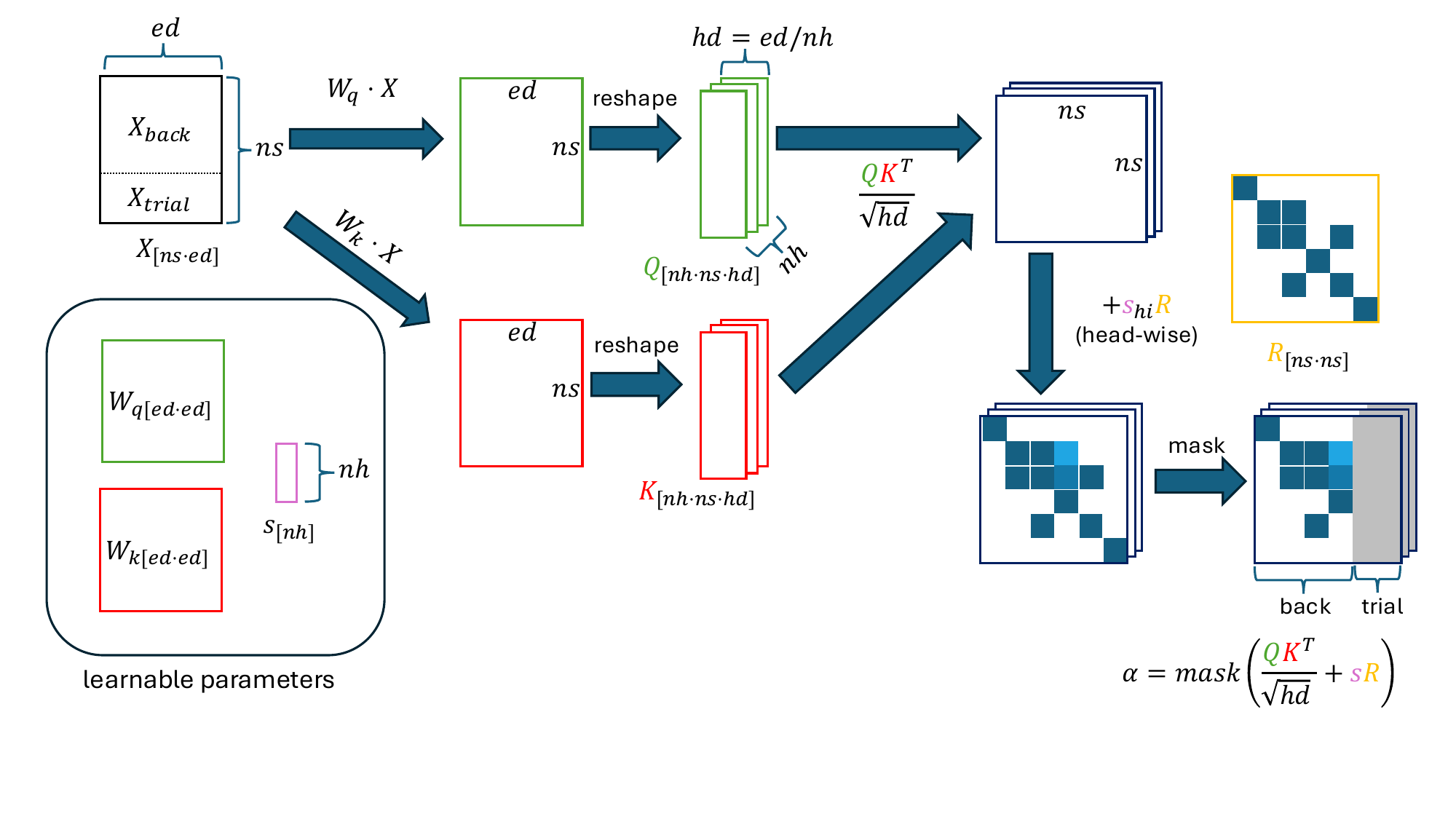} 
	\caption{Relational Multi-Head Attention: attention map $ \alpha $ constructed from the input matrix $ X $ and the relationship matrix $ R $ }
	\label{fig:rel_mha}
\end{figure}

Figure \ref{fig:rel_mha} demonstrates the construction of relational attention map for our custom Relational Multi-Head Attention (MHA).
We use the following notation:
\begin{itemize}
	\item $ ns $ -- number of samples (trial + background)
	\item $ ed $ -- embedding dimension (number of features in the projection space)
	\item $ nh $ -- number of attention heads
	\item $ hd = \frac{ed}{nh} $ -- single head dimension
	\item $ R_{[ns \cdot ns]} $ -- $ ns \times ns $ symmetric relationships matrix
	\item $ W_q $, $ W_k $ -- queries and keys weights matrices, respectively
	\item $ s $ -- a learnable vector of length $ nh $ -- $ R $ scales for each attention head
\end{itemize}
Given a matrix $ X $, consisting of background and trial samples, our attention module multiplies it by queries and keys matrices $ W_q $ and $ W_k $, splits by head, and multiplies queries by keys head-wise.
The resulting matrices of size $ n_{samples} \times n_{samples} $ are then summed with the relationship matrix $ R $ multiplied by the learnable weight $ s_{hi} $, where $ hi$ is the head index.
The attention map $ \alpha $ is the result of masking the last $ n_{trial} $ columns in each matrix (this operation ensures that trial samples do not "attend" one another).
After the softmax operation and dropout, the attention maps (which now become attention weights) are multiplied by the values $ V $ ($ V = W_v \times X $, $ W_v $ is a learnable matrix).

Note that each row of $ X $ contains both features $ x_i $ and responses $ y_i $, but for trial samples $ y_i $ is set to zero because it is the prediction target.
During training, the data set is split into three sub-samples: trials, background, validation; both the validation and trials response variable values are not inputted to the transformer (set to zero), and the loss function is calculated by summing over the predictions for trials; validation samples are used as trial in each attention module.

